\documentclass{article}
\usepackage{spconf,amsmath,graphicx}
\usepackage{booktabs}

\title{FRNET: Flattened Residual Network for Infant MRI Skull Stripping}
\name{Qian Zhang$^{1,2}$, Li Wang$^2$, Xiaopeng Zong$^2$, Weili Lin$^2$, Gang Li$^2$, Dinggang Shen$^2$\thanks{This work is supported in part by NIH grants (MH100217, MH107815, MH108914, MH109773, MH116225, MH117943, HD053000, MH104324 and U01MH110274) and the efforts of the UNC/UMN Baby Connectome Project Consortium.}}

\address{$^1$Department of Computer Science, University of North Carolina at Chapel Hill, Chapel Hill, USA \\ $^2$Department of Radiology and BRIC, University of North Carolina at Chapel Hill, Chapel Hill, USA }

\begin{document}
%
\maketitle
\begin{abstract}
Skull stripping for brain MR images is a basic segmentation task. Although many methods have been proposed, most of them focused mainly on the adult MR images. Skull stripping for infant MR images is more challenging due to the small size and dynamic intensity changes of brain tissues during the early ages. In this paper, we propose a novel CNN based framework to robustly extract brain region from infant MR image without any human assistance. Specifically, we propose a simplified but more robust flattened residual network architecture (FRnet). We also introduce a new boundary loss function to highlight ambiguous and low contrast regions between brain and non-brain regions. To make the whole framework more robust to MR images with different imaging quality, we further introduce an artifact simulator for data augmentation. We have trained and tested our proposed framework on a large dataset (N=343), covering newborns to 48-month-olds, and obtained performance better than the state-of-the-art methods in all age groups. 
\end{abstract}
\section{Introduction}
\label{sec:intro}

\begin{figure}[htb]

\begin{minipage}[b]{.30\linewidth}
  \centering
  \centerline{\includegraphics[width=2.6cm]{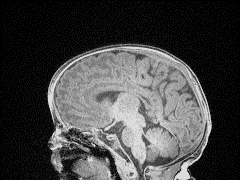}}
  \centerline{newborn}\medskip
\end{minipage}
\hfill
\begin{minipage}[b]{0.30\linewidth}
  \centering
  \centerline{\includegraphics[width=2.6cm]{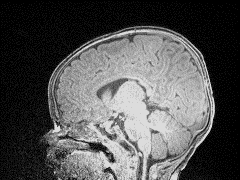}}
  \centerline{3 months}\medskip
\end{minipage}
\hfill
\begin{minipage}[b]{0.30\linewidth}
  \centering
  \centerline{\includegraphics[width=2.6cm]{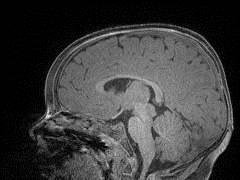}}
  \centerline{6 months}\medskip
\end{minipage}

\begin{minipage}[b]{.30\linewidth}
  \centering
  \centerline{\includegraphics[width=2.6cm]{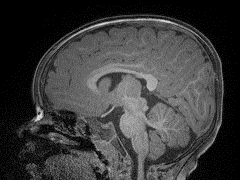}}
  \centerline{12 months}\medskip
\end{minipage}
\hfill
\begin{minipage}[b]{0.30\linewidth}
  \centering
  \centerline{\includegraphics[width=2.6cm]{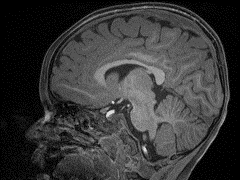}}
  \centerline{24 months}\medskip
\end{minipage}
\hfill
\begin{minipage}[b]{0.30\linewidth}
  \centering
  \centerline{\includegraphics[width=2.6cm]{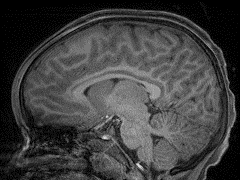}}
  \centerline{48 months}\medskip
\end{minipage}
\caption{T1-weighted infant brain MR images in the first 4 years of life. As can be seen, image contrast is low, and also imaging intensity, brain size and shape change dynamically.}
\label{fig:ages}
\end{figure}

Skull stripping, also called brain extraction, aims to retain brain parenchyema and discard non-brain tissues, such as skull, scalp, and dura \cite{c1}. As a fundamental problem in brain MR image analysis, numerous methods have been proposed over the past 20 years. Some of them are based on morphological operations, e.g., brain surface extraction (BSE) \cite{c2} and some others are based on deformation models that try to fit the brain surface, e.g., the brain extraction tool (BET) \cite{c1}. However, most of these methods only focus on adult MR images, and there are only a few methods dedicated on infant brain MRI \cite{c3,c4}. The main challenge for skull stripping of infant brain MR images is the rapid change of brain tissues during the early life period \cite{c5}. As an example, \textbf{Fig.~\ref{fig:ages}} shows infant brain images in the first 4 years of life. We can see low contrast, and dynamic changes of imaging intensity, brain size and shape in these images. Recently, deep convolutional neural networks (CNNs) \cite{c6} have achieved great success in medical image segmentation. Among them, UNet \cite{c7}, an evolutionary variant of CNN, has achieved excellent performance by effectively combining upper-level features and low-level features in the network architecture. Inspired by UNet, in this paper, we propose a new deep learning-based framework to deal with size variance and dynamic intensity changes of MR images in different age groups.
In the training process, we introduce a new boundary loss function, by making use of spatial information and applying voxel-wise weights to improve the training speed and avoid local minimum. As shown in \textbf{Fig.~\ref{fig:results}}, this loss function can largely increase the segmentation accuracy. We have also introduced an artifact simulator for data augmentation to make our proposed framework more robust to low-quality images.

\section{METHOD}

\subsection{Network Architecture}

\begin{figure}[htb]

\begin{minipage}[b]{1.0\linewidth}
  \centering
  \centerline{\includegraphics[width=8.5cm]{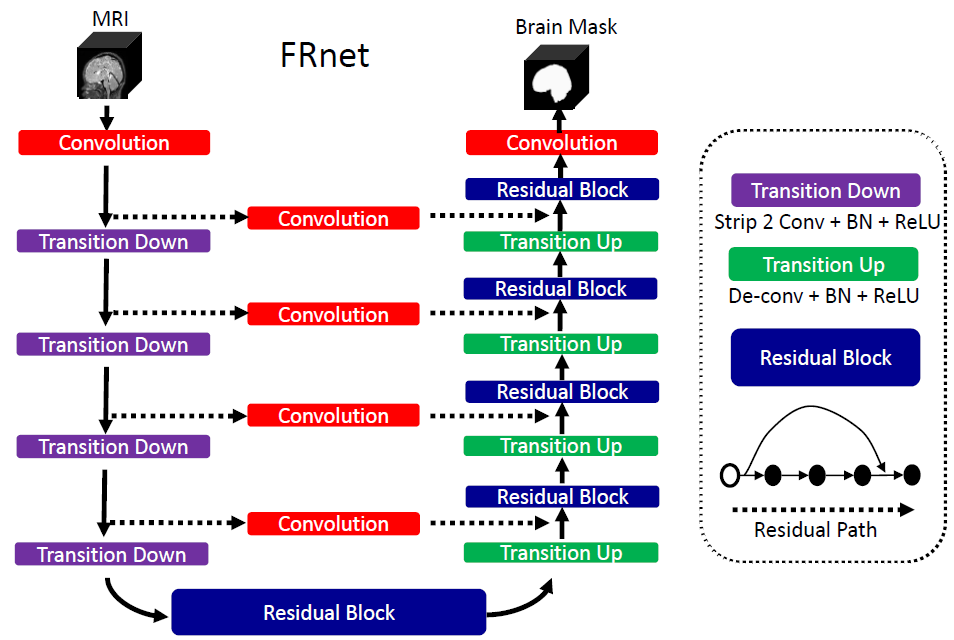}}
\end{minipage}
\caption{The architecture of the proposed flattened residual net (FRnet). The left portion is a successive set of down-sampling layers and the right portion is a combination of convolutional layers and up-sampling layers. We use convolutional and de-convolutional layers with 2x2x2 kernel and stride 2 for both down-sampling and up-sampling.}
\label{fig:frnet}
\end{figure}

Our proposed flattened residual net (FRnet) is motivated from UNet \cite{c7}. To address the over-fitting issue of UNet on some datasets and enhance its generalization ability to adapt image changes in different age groups: 1) we simplify the encoder section and apply only one convolution after down-sampling in each layer; 2) we use strip 2 convolutional layers for down-sampling and also the deconvolutional layers for up-sampling; 3) we introduce some residual paths in the decoder section to help training. The architecture of FRnet is shown in \textbf{Fig.~\ref{fig:frnet}}.

\vspace{-0.3cm}
\subsection{Boundary Loss}

The output of the network goes through a soft-max layer, with the value of each voxel representing the probability for each class. In most studies, people use cross-entropy loss to train the network. However, it is very common that this loss only decreases for a while at the beginning of the training process, and then shakes harshly and ends up at certain local minima. This yields a network that can only roughly point out the locations of the regions but fail to produce the details. Some methods have been proposed to fix this issue by using weighted loss functions i.e., weighted cross-entropy (WCE) loss \cite{c8}, which applies a weight for each class during the training process as shown below.

\vspace{-0.3cm}
\begin{eqnarray}
WCE(p_t)=-\alpha log(p_t)
\end{eqnarray}

\noindent where $p_t$ is the predicted probability of the targeted voxel, and $\alpha$ is a manually selected weight for each class. Usually the loss value of this function should also be divided by the volume size during the training.

An improved version of WCE is the well-known focal loss \cite{c9}, which modifies the weight dynamically according to the output of the network:

\vspace{-0.2cm}
\begin{eqnarray}
FL(p_t)=-(1-p_t)^\gamma log(p_t)
\end{eqnarray}

Our proposed method, boundary loss, takes into consideration the spatial information derived from the whole brain mask. To avoid being dominated by the easily classified inner regions in the training process, we increase the weight of loss generated by the voxels near the boundary during the training process.

\begin{figure}[htb]

\begin{minipage}[b]{.30\linewidth}
  \centering
  \centerline{\includegraphics[width=2.6cm]{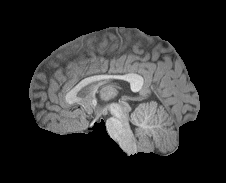}}
\end{minipage}
\hfill
\begin{minipage}[b]{0.30\linewidth}
  \centering
  \centerline{\includegraphics[width=2.6cm]{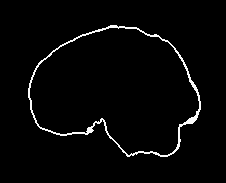}}
\end{minipage}
\hfill
\begin{minipage}[b]{0.30\linewidth}
  \centering
  \centerline{\includegraphics[width=2.6cm]{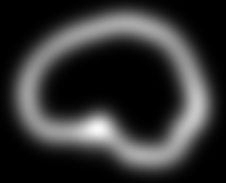}}
\end{minipage}
\caption{The construction of the density map $B$. (From left to right: the skull stripped image, the brain boundary and the density map.}
\label{fig:bloss}
\end{figure}

\vspace{-0.5cm}
\begin{eqnarray}
BL(p_t)=-B log(p_t)
\end{eqnarray}

\noindent where $B$ is a density map derived from the boundary of our targeted region using Gaussian filter as shown in \textbf{Fig.~\ref{fig:bloss}}.

\subsection{Artifact Simulator}

MR Imaging is a time-consuming process, so it is very sensitive to motion, which can cause ghosting, blurring and geometric distortion in the images and heavily corrupt further analysis of the image. Over the past 30 years since MRI has been used for medical diagnosis, numerous methods have been proposed to mitigate or to correct this artifact, but still, there is no single method that can be applied in all imaging situations \cite{c10}, due to the huge number of unknown parameters of the motion. Therefore, instead of trying to correct the motion artifacts, we propose to use an artifact simulator for data augmentation during the training process, for improving our simple task of skull-stripping, not image acquisition. We find that this method can also improve the robustness of the whole skull-stripping framework, for both the images with and without artifacts.

\begin{figure}[htb]

\begin{minipage}[b]{.30\linewidth}
  \centering
  \centerline{\includegraphics[width=2.6cm]{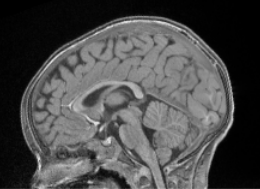}}
\end{minipage}
\hfill
\begin{minipage}[b]{0.30\linewidth}
  \centering
  \centerline{\includegraphics[width=2.6cm]{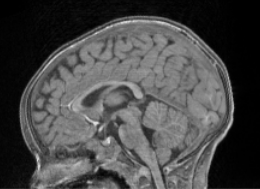}}
\end{minipage}
\hfill
\begin{minipage}[b]{0.30\linewidth}
  \centering
  \centerline{\includegraphics[width=2.6cm]{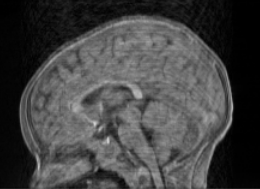}}
\end{minipage}
\caption{Image produced by our artifact simulator. (From left to right: the original image, the image with artifact using $\sigma = 0.3$ and the image with artifact using $\sigma = 1.0$).}
\label{fig:artifact}
\end{figure}

The raw data acquired by a MR scanner are stored in the so-called k-space domain \cite{c11}. To simulate the motion affects in a MR image, we first use a Fourier transformation to obtain the k-space data. Then we apply a random phase shift to each k-space line along the readout direction, where the random phase shift is given by:

\vspace{-0.3cm}
\begin{eqnarray}
exp(i\cdot k_y \cdot \sigma _y + i \cdot k_z \cdot \sigma _z)
\end{eqnarray}

\noindent where $k_y$ and $k_z$ are the k-space coordinates along the phase and partition encoding directions for the k-space line, and $\sigma _y$ and $\sigma _z$ are the random motion amounts.  The random motion amounts are assumed to follow a Gaussian distribution. After applying the random phase shift, inverse Fourier transform is applied to the k-space data to obtain the motion corrupted images. The standard deviation $\sigma$ of the Gaussian distribution is varied to generate images with different severities of motion artifacts. \textbf{Fig.~\ref{fig:artifact}} shows the simulated images.

\vspace{-0.2cm}
\section{EXPERIMENTS}
\vspace{-0.3cm}
\subsection{Dataset}
All data used in this study were obtained from UNC/UMN Baby Connectome Project (BCP) dataset \cite{c12}. All scans for subjects less than two years old were acquired while the infants were naturally sleeping and fitted with ear protection, with their heads secured in a vacuum-fixation device. T1w MR images were acquired with 320 sagittal slices using parameters: TR/TE=2400/2.24 $ms$ and resolution=0.8x0.8x0.8 $mm^3$. 

As shown in \textbf{Table.~\ref{table:1}}, all scans were labeled by human experts and were grouped into 7 classes:  0~(newborn), 3, 6, 9, 12, 24, 48 months of age. We randomly chose 20\% from each class as our testing set.

\vspace{-0.5cm}
\begin{table}[htb]
\caption{The number of scans in each age group.}
\begin{center}
\begin{tabular}{lcccccccccl}
\toprule
Months & 0 & 3 & 6 & 9 & 12 & 24 & 48 & Total \\
\midrule
Training & 3 & 38 & 47 & 53 & 59 & 35 & 43 & 278  \\
Testing & 2 & 6 & 12 & 13 & 14 & 8 & 10 & 65 \\
\bottomrule
\end{tabular}
\end{center}
\label{table:1}
\end{table}

\vspace{-0.7cm}
\subsection{Implementation Details}
We chose pytorch to implement our network. We also used python libraries SimpleITK and Scikit-image for data preprocessing. We trained and tested our framework on a Linux workstation equipped with an Intel Xeon E5-2650 v4 CPU and 12 GB NVIDIA TITAN Xp GPUs.

All MR images were preprocessed with N4 bias correction \cite{tustison2010n4itk}, and we also used randomized reorientation and resizing for data augmentation. For each training image, we used the artifact simulator to generate two simulated images for training with $\sigma$ equals 0.3 and 1.0 respectively. 

All the CNN models were initialized with xavier \cite{c13}, and we chose Adam as the optimizer with a fixed learning rate of 0.003. We used a 2-fold cross-validation on the training set to select the best model for testing.

\subsection{Results}
\begin{figure}[hbt]

  \centering
  \centerline{\includegraphics[width=8cm]{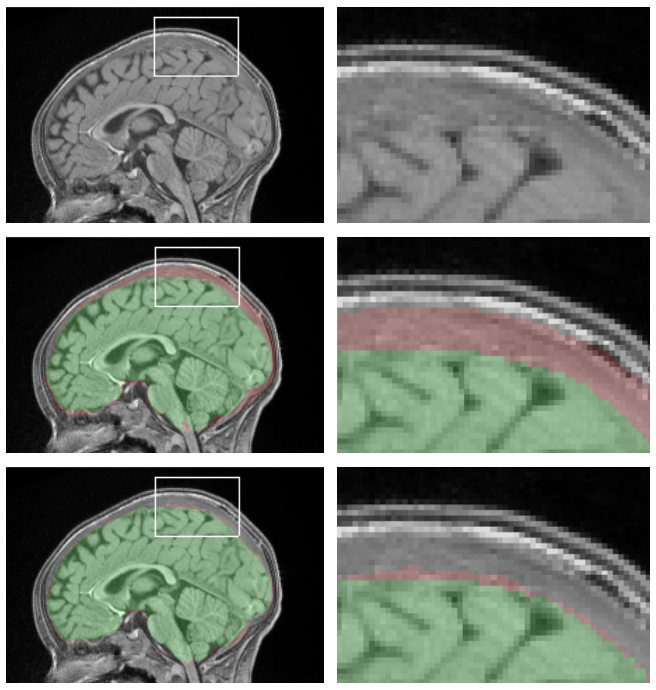}}
\caption{From top to bottom are the original image, the results of cross-entropy loss and the results of our proposed boundary loss. Correctly segmented regions are marked with green, and incorrectly segmented regions are marked with red.}
\label{fig:results}
\end{figure}

As shown in \textbf{Fig.~\ref{fig:results}}, the widely used cross-entropy loss func- tion cannot clearly distinguish the brain and non-brain regions because the inner easy-to-segment region is too large and stops the training process. On the contrary, our proposed boundary loss forces the optimizer to focus more on the outer regions and thus produces a much better result.

We further quantitatively evaluate our two main contributions: 1) the proposed FRnet architecture for segmentation and 2) the boundary loss for training. For the traditional methods, we chose BSE and BET for comparison, while, for deep learning-based methods, we chose UNet. We also tested different loss functions on each network architecture. The results are shown in  \textbf{Table.~2}.

\begin{table*}
\newcommand{\tabincell}[2]{\begin{tabular}{@{}#1@{}}#2\end{tabular}}
\centering 
\begin{tabular}{ccccccccccl}
\toprule
Month & 0 & 3 & 6 & 9 & 
12 & 24 & 48 & Overall \\
\midrule
BSE 
& \tabincell{c}{0.893\\$\pm$0.035} 
& \tabincell{c}{0.930\\$\pm$0.013}
& \tabincell{c}{0.947\\$\pm$0.007} 
& \tabincell{c}{0.950\\$\pm$0.010} 
& \tabincell{c}{0.954\\$\pm$0.003} 
& \tabincell{c}{0.951\\$\pm$0.009} 
& \tabincell{c}{0.957\\$\pm$0.012} 
& \tabincell{c}{0.941\\$\pm$0.027}  \\
\hline
BET
& \tabincell{c}{0.897\\$\pm$0.072} 
& \tabincell{c}{0.929\\$\pm$0.033}
& \tabincell{c}{0.955\\$\pm$0.012} 
& \tabincell{c}{0.960\\$\pm$0.011} 
& \tabincell{c}{0.960\\$\pm$0.011} 
& \tabincell{c}{0.971\\$\pm$0.004} 
& \tabincell{c}{0.976\\$\pm$0.003} 
& \tabincell{c}{0.951\\$\pm$0.141}  \\
\hline
\tabincell{c}{UNet\\(Cross-entropy loss)} 
& \tabincell{c}{0.922\\$\pm$0.070} 
& \tabincell{c}{0.923\\$\pm$0.072}
& \tabincell{c}{0.936\\$\pm$0.045} 
& \tabincell{c}{0.911\\$\pm$0.053} 
& \tabincell{c}{0.930\\$\pm$0.057} 
& \tabincell{c}{0.959\\$\pm$0.027} 
& \tabincell{c}{0.938\\$\pm$0.060} 
& \tabincell{c}{0.931\\$\pm$0.060}  \\
\hline
\tabincell{c}{UNet\\(Boundary loss)} 
& \tabincell{c}{0.957\\$\pm$0.076} 
& \tabincell{c}{0.954\\$\pm$0.064}
& \tabincell{c}{0.972\\$\pm$0.025} 
& \tabincell{c}{\textbf{0.986}\\$\pm$0.002} 
& \tabincell{c}{0.972\\$\pm$0.025} 
& \tabincell{c}{0.919\\$\pm$0.087} 
& \tabincell{c}{0.970\\$\pm$0.047} 
& \tabincell{c}{0.964\\$\pm$0.056}  \\
\hline
\tabincell{c}{FRnet\\(WCE loss)} 
& \tabincell{c}{0.963\\$\pm$0.061} 
& \tabincell{c}{0.961\\$\pm$0.056}
& \tabincell{c}{0.979\\$\pm$0.008} 
& \tabincell{c}{0.982\\$\pm$0.004} 
& \tabincell{c}{0.972\\$\pm$0.014} 
& \tabincell{c}{0.982\\$\pm$0.003} 
& \tabincell{c}{0.970\\$\pm$0.042} 
& \tabincell{c}{0.972\\$\pm$0.040}  \\
\hline
\tabincell{c}{FRnet\\(Focal loss)} 
& \tabincell{c}{0.979\\$\pm$0.007} 
& \tabincell{c}{0.980\\$\pm$0.007}
& \tabincell{c}{0.981\\$\pm$0.003} 
& \tabincell{c}{0.982\\$\pm$0.002} 
& \tabincell{c}{0.979\\$\pm$0.006} 
& \tabincell{c}{0.980\\$\pm$0.003} 
& \tabincell{c}{0.982\\$\pm$0.005} 
& \tabincell{c}{0.980\\$\pm$0.006}  \\
\hline
\tabincell{c}{FRnet\\(Boundary loss)} 
& \tabincell{c}{\textbf{0.985}\\$\pm$0.004} 
& \tabincell{c}{\textbf{0.986}\\$\pm$0.003}
& \tabincell{c}{\textbf{0.986}\\$\pm$0.002} 
& \tabincell{c}{\textbf{0.986}\\$\pm$0.002} 
& \tabincell{c}{\textbf{0.984}\\$\pm$0.005} 
& \tabincell{c}{\textbf{0.985}\\$\pm$0.002} 
& \tabincell{c}{\textbf{0.987}\\$\pm$0.004} 
& \tabincell{c}{\textbf{0.986}\\$\pm$0.003}  \\
\bottomrule
\end{tabular}
\\
\raggedright
\textbf{Table.~2} The mean Dice scores on different age groups for all methods. First, we can find that conventional
methods that focus on adult brains all have low performance on early aged MR images, while deep learning-based
methods generally are more adaptive to images from different age groups. More specifically, our proposed FRnet
outperforms UNet and has much smaller standard deviation, we believe this is because the use of the simplified
encoder sections in FRnet, which makes it more robust to intensity difference of different age groups.
\label{table:2}
\end{table*}

\section{CONCLUSION}
\vspace{-0.1cm}
In this work, we have proposed a novel framework to train a CNN model for skull stripping of infant brain MR images. We believe the same framework could also be applied for other segmentation tasks like the white matter (WM) and gray matter (GM) segmentation. Our proposed FRnet is a simplified version of UNet,and we have demonstrated its robustness to dynamic intensity changes and low contrast as shown in MR images of early age groups. Since UNet is a widely used deep learning model for segmentation tasks, we believe our FRnet can help related applications where UNet is used. Also, our proposed boundary loss would also benefit other tasks because the boundary is always the most crucial part in segmentation map. The artifact simulator could also be helpful in some other tasks such as registration and landmark detection, because motion artifact is a common problem in MR related applications.

\vspace{-0.4cm}
\bibliographystyle{IEEEbib}
\bibliography{strings,refs}

\end{document}